%% file: templateArxiv.tex
\documentclass{article}

\usepackage{PRIMEarxiv}

\usepackage[utf8]{inputenc} 
\usepackage[T1]{fontenc}    
\usepackage{hyperref}       
\usepackage{url}            
\usepackage{booktabs}       
\usepackage{amsfonts}       
\usepackage{nicefrac}       
\usepackage{microtype}      
\usepackage{lipsum}
\usepackage{fancyhdr}       
\usepackage{graphicx}       
\graphicspath{{media/}}     

\usepackage{algorithm}
\usepackage{algorithmic}
\usepackage{microtype}
\usepackage{graphicx}
\usepackage{subcaption}
\usepackage{booktabs} 
\usepackage{multirow}
\usepackage{enumitem}

\def \etal {{et al.\thinspace}}
\newcommand{\eg}{\emph{e.g.}}

\newcommand{\myparagraph}[1]{\noindent\textbf{#1}}

\def \sys {InvLBA}

\usepackage{amsmath}

\usepackage{amsmath}
\usepackage{amssymb}
\usepackage{mathtools}
\usepackage{amsthm}
\usepackage{xcolor}
\usepackage{arydshln}

\usepackage[capitalize,noabbrev]{cleveref}
\theoremstyle{plain}
\newtheorem{theorem}{Theorem}[section]

\theoremstyle{definition}

\theoremstyle{remark}
\newtheorem{remark}[theorem]{Remark}

\usepackage[textsize=tiny]{todonotes}

\title{Invisible Clean-Label Backdoor Attacks for Generative Data Augmentation
}

\author{
  Ting Xiang, Jinhui Zhao, Changjian Chen, Zhuo Tang \\
  Hunan University \\
  \texttt{\{txiang, zachary, changjianchen, ztang\}@hnu.edu.cn} \\
}

\begin{document}
\maketitle

\input{sec/0_abstract}


\input{sec/1_intro}
\input{sec/2_related_work}
\input{sec/3_formulation}
\input{sec/4_theoretical}

\input{sec/5_method}
\input{sec/6_experiment}

\input{sec/7_conclusion}


\bibliographystyle{unsrt}  
\bibliography{references}  

\newpage
\appendix
\input{sec/8_appendix}

\end{document}

%% file: sec/0_abstract.tex
\begin{abstract}
With the rapid advancement of image generative models, generative data augmentation has become an effective way to enrich training images, especially when only small-scale datasets are available. At the same time, in practical applications, generative data augmentation can be vulnerable to clean-label backdoor attacks, which aim to bypass human inspection. However, based on theoretical analysis and preliminary experiments, we observe that directly applying existing pixel-level clean-label backdoor attack methods (\eg, COMBAT) to generated images results in low attack success rates. This motivates us to move beyond pixel-level triggers and focus instead on the latent feature level.
To this end, we propose InvLBA, an invisible clean-label backdoor attack method for generative data augmentation by latent perturbation. We theoretically prove that the generalization of the clean accuracy and attack success rates of InvLBA can be guaranteed. Experiments on multiple datasets show that our method improves the attack success rate by 46.43\% on average, with almost no reduction in clean accuracy and high robustness against SOTA defense methods.
  
\end{abstract}

%% file: sec/1_intro.tex
\section{Introduction}

The success of deep learning fundamentally relies on large-scale, high-quality training data~\cite{xiang2025enhancing}. 
However, manually collecting such datasets is prohibitively expensive and time-consuming.
Recently, with advances in generative models, such as diffusion models~\cite{rombach2022high}, generative data augmentation~(GDA) has emerged as a promising solution to this challenge, such as DistDiff~\cite{zhu2024distribution}.
However, this new paradigm of data acquisition also introduces vulnerabilities to clean-label backdoor attacks, in which attackers poison generated data with invisible triggers while keeping labels unchanged.
Models trained on such poisoned datasets can still achieve strong performance on clean data, while producing attacker-specified predictions when the trigger is present~\cite{gu2017badnets}.
Such vulnerabilities in GDA are more severe than in natural-image settings because the unavoidable artifacts in generated images make triggers harder to identify.
Therefore, understanding security risks in GDA is of critical importance.
We hope our work highlights the potential security risks of GDA,
and calls for further attention to the robustness and safety of models trained on generated data.
To the best of our knowledge, there is no existing method specified for the invisible clean-label backdoor attacks in GDA.

To perform invisible clean-label backdoor attacks in GDA, a straightforward way is to apply a clean-label attack method (\eg, COMBAT~\cite{huynh2024combat}) to the generated images.
However, our initial attempts have shown that existing clean-label backdoor attack methods on generated images lead to a large drop in attack success rates compared to natural images.
To understand why they fail in GDA, we conduct a theoretical analysis, which shows that a higher adversarial perturbation sensitivity~(APS) leads to lower attack success rates, thereby explaining the attack success rate degradation of existing clean-label methods in GDA.
Additionally, we conduct an experiment to validate our theoretical analysis, demonstrating that APS increases on the generated images.



Motivated by the theoretical analysis, we propose to apply latent triggers, which yield lower APS.
To this end, we propose {\sys}, an invisible clean-label backdoor attack method for generative data augmentation with latent perturbation.
{\sys} incorporates backdoor injection directly into the generation process of the Stable Diffusion model, with a one-step noise prediction strategy to enable an end-to-end training.
We theoretically prove that the generalization error of clean images for our method will approach zero as the dataset size increases, even in the presence of triggers.
Moreover, the generalization error of poisoned images is upper-bounded with a small value.
We evaluate our method on a diverse set of benchmark datasets widely used in both GDA and backdoor attack research.
Extensive experiments demonstrate that {\sys} is highly effective and achieves strong generalization across model architectures.
Defense experiments on state-of-the-art~(SOTA) defense methods also show that our method is robust against these defenses.

In summary, our main contributions are threefold:
\begin{itemize}[itemsep=2pt, topsep=0pt, parsep=0pt]
     \item \textbf{A theoretical analysis} for explaining why existing clean-label attack methods fail in generated images.
    
    \item \textbf{An invisible clean-label backdoor attack for GDA} with theoretical guarantees on the generalization of clean accuracy and attack success rates.
    
    \item \textbf{A series of experimental results} 
    demonstrate the effectiveness, generalization, and robustness against SOTA defenses of our method.
\end{itemize}



%% file: sec/2_related_work.tex
\section{Related Work}
Since our method focuses on backdoor attacks arising in GDA pipelines, rather than on improving the GDA methods themselves, we limit our review to prior studies on backdoor attacks and corresponding defense techniques.

\subsection{Backdoor Attack}
A backdoor attack~\cite{gu2017badnets} refers to a training-time poisoning attack in which an attacker injects a trigger pattern into a subset of training samples to manipulate the learned model behavior.
The resulting model, also called a victim model, behaves normally on clean inputs, while predicting an attacker-specified target class when the trigger is present.
Based on whether the labels of poisoned samples are modified, backdoor attacks can be classified into two categories: dirty-label attacks and clean-label attacks.

\myparagraph{Dirty-label attacks} inject a trigger into a subset of non-target-class samples and explicitly modify their label to the pre-defined target label.
The pioneering work along this line was proposed by Gu~\etal~\cite{gu2017badnets}, which has demonstrated that a fixed image patch can be used as a trigger and reliably associated with the target class.
Although simple and effective, such patch-based triggers are visually salient and can be easily identified by human inspection.
Subsequent studies improved trigger stealthiness through various designs, including crafting a warping-based trigger~\cite{nguyenwanet}, training trigger generators~\cite{nguyen2020input, doan2021lira}, optimizing feature-space perturbations to make the trigger imperceptible~\cite{doan2021backdoor, ZhongQZ22Imperceptible}, and injecting the trigger into the frequency domain~\cite{wang2021backdoor, hammoud2021check, feng2022fiba}.
Despite the improved visual stealthiness, dirty-label backdoor attacks remain susceptible to detection, as the poisoned samples exhibit semantic inconsistencies with their modified target labels.

\myparagraph{Clean-label attacks} address the above limitation by injecting a trigger into a subset of target-class samples without modifying their ground-truth labels. 
The pioneering work proposed by \cite{turner2019label} observed that naively applying a fixed trigger to target-class samples leads to ineffective backdoor attacks, 
thus they proposed to perturb target-class samples to make them difficult to classify, forcing the model to rely on the injected trigger when learning the target class.
Subsequent studies further improved clean-label attacks by optimizing a trade-off between attack effectiveness and trigger stealthiness. 
Barni~\etal\cite{barni2019new} and Liu~\etal\cite{liu2020reflection} enhanced stealthiness by using sinusoidal stripe patterns and image reflection as triggers, respectively.
Hidden Trigger Backdoor Attacks~\cite{saha2020hidden} proposed to optimize poisoned samples to remain close to target-class samples in pixel space,  
while simultaneously matching the feature representations  
of non-target samples injected with the trigger.  
Sleeper Agent~\cite{souri2022sleeper} further extended this line of work by applying a gradient-based optimization strategy to explicitly enforce feature-space alignment.
More recently, Narcissus~\cite{zeng2023narcissus} and COMBAT~\cite{huynh2024combat} improved attack effectiveness 
by optimizing the trigger or trigger generator through training a surrogate downstream victim model.

Building upon existing invisible clean-label backdoor attacks, our work focuses on backdoor attacks in the context of GDA, 
where training datasets consist of both original data and generated data produced by generative models.
For this purpose, we analyze the feasibility of directly applying existing clean-label attack methods to the GDA pipeline, and propose InvLBA, an invisible clean-label backdoor attack method that injects triggers in the latent space of a Stable Diffusion model.

\subsection{Backdoor Defense}
To mitigate the security risks introduced by backdoor attacks, a wide range of defense methods have been proposed.
According to the defense stage applied, existing methods can be broadly categorized into data-level defenses and model-level defenses.

\myparagraph{Data-level defenses} aim to detect and filter out poisoned samples, at either the training or testing stage.
Based on the stage at which they are applied,  
these methods can be categorized into training data defenses and test data defenses.
Training data defenses focus on detecting and purifying poisoned training samples through feature-based analysis.
Representative methods such as spectral signature analysis~\cite{tran2018spectral}, activation clustering patterns~\cite{chen2018detecting}, amplified corrupted feature identified through robust covariance estimation~\cite{hayase2021spectre} and diffusion latent features analysis~\cite{zhou2024dataelixir}.
Test data defenses focus on filtering out malicious inputs during model inference using different detection strategies.
For example, STRIP~\cite{gao2019strip} detected poisoned test inputs based on their anomalously low prediction entropy, and Februus~\cite{doan2020februus} leveraged GradCAM to localize and remove suspicious trigger regions.

\myparagraph{Model-level defenses} aim to detect poisoned models or mitigate backdoor effects within compromised models.
Based on the objective, existing methods can be categorized into detection-based defenses and mitigation-based defenses.
Detection-based defenses analyze model responses under clean inputs to identify the presence of the backdoor.
Representative examples include optimal class-inducing patterns~\cite{wang2019neural}, scanned neurons~\cite{liu2019abs} and optimized universal litmus patterns~\cite{kolouri2020universal}.
Mitigation-based defenses aim to modify the victim model to suppress or remove backdoor behaviors.
Such as, Fine-pruning~\cite{liu2018fine} removed neurons that are inactive on clean data,  
and Bridging Mode Connectivity~\cite{zhao2020bridging} leveraged mode connectivity to mitigate backdoor effects in the model.

%% file: sec/3_formulation.tex
\section{Problem Formulation}
\label{Sec:problem_formulation}
In this section, we first briefly introduce the clean-label backdoor attack problem and formalize how it arises in the context of generative data augmentation~(GDA).

\myparagraph{Clean-label backdoor attack problem.} 
We are given $n$ original training images and their labels $\mathbf{D}_o=\{(x_1, y_1), \dots, (x_n, y_n)\}$ sampled from a distribution $\mathcal{D}$, where $x \in \mathcal{X} \in \mathbb{R}^D$, $y \in \mathcal{Y} = \{1, \dots, C\}$,
$D$ is the input dimension, $C$ is the number of classes. 
The dataset can be decomposed as $\mathbf{D}_o = \bigcup_{j=1}^C \mathbf{D}_o^j$, where $\mathbf{D}_o^j$ is the subset of samples belonging to class $j$. 

We consider a clean-label backdoor attack on target class $c \in \mathcal{Y}$, against a classification model $f(x;\theta):\{\mathcal{X},\Theta\} \rightarrow \mathcal{Y}$ parameterized by $\theta$.
Given a poison rate $\alpha$, the attacker samples a subset $\mathbf{S}_o^c \subseteq \mathbf{D}_o^c$, such that $\alpha=|\mathbf{S}_o^c|/|\mathbf{D}_o^c|$.
A trigger $P(\cdot)$ is then injected to data in $\mathbf{S}_o^c$, yielding a poisoned subset $\mathbf{S}_p^c$.
To ensure the stealth of the attack, the trigger $P(\cdot)$ must be perceptually invisible for humans.
All remaining samples in $\mathbf{D}_o$ keep unchanged, forming the clean subset $\mathbf{D}_o'=\mathbf{D}_o \setminus \mathbf{S}_o^c$.
The combined set $\mathbf{D}_p = \mathbf{S}_p^c \cup \mathbf{D}_o'$ forms the final poisoned training dataset.
The victim model is trained by minimizing the empirical risk:
\begin{equation}
\label{Eq:vic}
    \theta = \arg \min_{\theta \in \Theta} \sum_{(x,y) \in \mathbf{D}_p} \mathcal{L} \big(f(x; \theta), y\big),
\end{equation}
where $\mathcal{L}$ refers to a classification loss function, such as cross-entropy loss.
The trained victim model $f$ is expected to archive minimum clean generalization error $\mathcal{E}_o(f, \mathcal{D})$ and poison generalization error $\mathcal{E}_p(f, \mathcal{D})$:
\begin{equation}
    \begin{aligned}
        & \mathcal{E}_o(f, \mathcal{D}) = \mathbb{E}_{(X,Y)\sim \mathcal{D}}[\mathcal{L}(f(X;\theta),Y)], \\
        & \mathcal{E}_p(f, \mathcal{D}) = \mathbb{E}_{(X,Y)\sim \mathcal{D}}[\mathcal{L}(f(X+P(X);\theta),c)]. \\
    \end{aligned}
\end{equation}


\myparagraph{Clean-label backdoor attacks for GDA.} 
Here we consider the scenario where clean-label backdoor attacks occur during GDA.
Given a GDA method $G(\cdot)$, such as DistDiff~\cite{zhu2024distribution}, for each original image $(x_i, y_i) \in \mathbf{D}_o$, $G(\cdot)$ is applied to generate $m$ augmented images.
We denote the set of generated images from $x_i$ as:
\begin{equation}
    \mathbf{G}_i  
    = \big\{  
    x_i^{k} \sim G(x_i)  
    \;\big|\;  
    k = 1, \dots, m  
    \big\},
\end{equation}
where $x_i^{k}$ denotes the $k$-th image generated from $x_i$, and $m$ is the expansion ratio.
The generated dataset is then constructed as:
\begin{equation}
    \mathbf{D}_g  
    = \bigcup_{i=1}^{n}  
    \big\{  
    (x, y_i)  
    \;\big|\;  
    x \in \mathbf{G}_i  
    \big\}.
\end{equation}
The generated dataset can also be decomposed as $\mathbf{D}_g = \bigcup_{j=1}^C \mathbf{D}_g^j$, where $\mathbf{D}_g^j$ is the subset of generated samples associated with class $j$.

The attacker then samples a subset $\mathbf{S}_g^c \subseteq \mathbf{D}_g^c$, such that poison rate $\alpha ={|\mathbf{S}_g^c|}/{|\mathbf{D}_g^c \cup \mathbf{D}_o^c|}$,
which corresponds to the fraction of poisoned samples among all target-class training data,  
including both original and generated samples.

Next, following the clean-label backdoor attack setting, $P(\cdot)$ is applied to data in $\mathbf{S}_g^c$, yielding the poisoned subset $\mathbf{S}_p^c$.
The clean generated subset $\mathbf{D}_g'=\mathbf{D}_g \setminus \mathbf{S}_g^c$, together with $\mathbf{D}_o$ and $\mathbf{S}_p^c$, construct the final poisoned training dataset $\mathbf{D}_p$:
\begin{equation}
    \mathbf{D}_p = \mathbf{D}_g' \cup \mathbf{D}_o \cup \mathbf{S}_p^c.
\end{equation}
The victim model is trained on $\mathbf{D}_p$ according to Eq.~(\ref{Eq:vic}).
The resulting victim model possesses the same characteristics as it does in clean-label backdoor attack scenarios.


%% file: sec/4_theoretical.tex
\section{Theoretical Analysis}
\label{sec:theoretical}

Our initial attempts show that applying existing pixel-level triggers directly on generated samples leads to a degradation in ASR.
To identify the root causes, we conduct a theoretical analysis.
From the work of Yu~\etal~\cite{yu2024generalization}, we have:
\begin{theorem}~\cite{yu2024generalization}
\label{thm:pge}
Let $N = |\mathbf{D}_p|$.
For any victim model $f$ and clean model $g \in \mathcal{H}_{W,D}$,
if the trigger $P(x)$ meets the following three conditions for some $\epsilon > 0$, $\tau > 0$, $\lambda \ge 1$: 
\begin{align*}
    \text{(c1):} \; & \mathbb{E}_{(x,y) \sim \mathcal{D}^{c}} \big[ g_c (x + P(x)) \big] \le \epsilon, \\
    \text{(c2):} \; & \mathbb{P}_{(x,y) \sim \mathcal{D}} \big(P(x) \in \mathcal{A} \mid y \neq c\big) \le \lambda \, \mathbb{P}_{(x,y) \sim \mathcal{D}} \big(P(x) \in \mathcal{A} \mid y = c\big), \; \text{for any set $\mathcal{A}$}, \\
    \text{(c3):} \; & \mathbb{E}_{x \sim \mathcal{D}}\Big[\big|(f-g)_{c} (P(x)) - (f-g)_{c}(x+P(x))\big|\Big] \le \tau, \; \text{where } (f-g)_{c}(x) = f_c(x) - g_c(x),
\end{align*}
where $f_c$ and $g_c$ denote the predicted probabilities of input x for class c. Then, with probability at least $1 - \delta - O(1/N)$, the poison generalization error holds for f:
\begin{align}
& \mathcal{E}_{p}(f, \mathcal{D})
\leq \lambda \cdot O \Bigg(
\frac{1}{\alpha}
\Big(
\mathbb{E}_{(x,y) \sim \mathbf{D}_p}[\mathcal{L}(f(x), y)]
 + \text{Rad}_{N}^{\mathcal{D}^{c}}(\mathcal{H})
\Big)
+ \sqrt{\frac{\ln(1/\delta)}{N\alpha}}
+ \epsilon
+ \tau
+ \frac{\lambda-1}{\lambda}
\Bigg).
\end{align}

\end{theorem}


In this theorem, apart from the empirical error (the first term) and the model complexity term (the second term), there are four additional terms that influence the upper bound. 
The residual term (the third term) depends solely on the size of the dataset, the poison rate $\alpha$ and $\delta$, which is independent of the data distribution (whether natural or generated). 
The term $\lambda$ is related only to the triggers and does not depend on the data distribution by definition. 
Moreover, as noted by Yu~\etal~\cite{yu2024generalization}, the term $\tau$ is influenced by the linear separability of the triggers, and is also independent of the data distribution.
Therefore, $\epsilon$ is the only term related to the data distribution, as it is clear that different images (whether natural or generated) combined with the triggers will lead to different outputs in the clean model.



The analysis above motivated us to hypothesize that $\epsilon$ is the root cause for the performance degradation of existing pixel-level attack methods.
To validate this, based on the definition of $\epsilon$, we introduce \textbf{Adversarial Perturbation Sensitivity (APS)}, denoted by $\mathcal{M}_{\mathcal{D}^{c}}$,
which is the expected prediction confidence of the clean model on the ground-truth label for target-class images after trigger injection:
\begin{equation}
\label{eq:APS}
    \mathcal{M}_{\mathcal{D}^{c}} = \mathbb{E}_{(x,y) \sim \mathcal{D}^{c}} \left[f_c(x+P(x)) \right],
\end{equation}
where $\mathcal{D}^{c}$ is the distribution of samples with the target label $c$.
Intuitively, when a trigger is applied, the lower prediction confidence of the target class on the clean model, the more likely images are attacked~\cite{turner2019label}.


To validate this, we calculated this metric separately for the original data and generated data. As shown in Table~\ref{tab:adversarial_perturbation_sensitivity},
by applying existing backdoor attack methods, original data exhibits lower APS, leading to a tighter upper bound on the poison generalization error, 
whereas generated data shows higher APS, yielding a looser upper bound.
More detailed proofs are provided in Appendix~\ref{pge}.

\begin{table}[t]
    \centering
    \caption{
    Comparison of APS across data sources and poisoning strategies.
    Original: poisoned natural data with existing pixel-level attacks;
    Generated: poisoned generated data with existing attacks;
    Generated (Ours): poisoned generated data with our method.
    }
    \resizebox{.6\linewidth}{!}{
    \label{tab:adversarial_perturbation_sensitivity}
    \renewcommand{\arraystretch}{1.2} 
    \begin{tabular}{@{}lccc@{}}
        \toprule
        \textbf{Dataset} & \textbf{Original} & \textbf{Generated} & \textbf{Generated (ours)} \\
        \midrule
        ImageNet-10-S       & 0.33 & 0.45 & 0.38 \\
        Pets             & 0.29 & 0.30 & 0.23\\
        \bottomrule
    \end{tabular}
    }
\end{table}

%% file: sec/5_method.tex
\section{Method} 

\begin{algorithm}[t]
\caption{The {\sys} Framework}
\label{alg:framework}
\begin{algorithmic}[]
\STATE \textbf{Input:} Original dataset $\mathbf{D}_o$, GDA method $G(\cdot)$, expansion ratio $m$, poison rate $\alpha$, target class label $c$, surrogate model $h$, max iterations $T$.
\STATE \textbf{Output:} Poisoned subset $\mathbf{S}_p^c$, clean expaned subset $\mathbf{D}_g'$, poisoned training dataset $\mathbf{D}_p$.
\STATE \hrulefill
\STATE \textbf{Stage 1: }Trigger training.
    \STATE Initialize $w \sim \mathcal{U}(0, 1)$, $b \sim \mathcal{N}(0, 1)$
    \FOR{$t=0 \to T-1$}
        \STATE $\{x, y\} \leftarrow SampleBatchOriginalData(\mathbf{D}_o)$
        \STATE Generate predicted images with trigger injection: $x'=\phi(x)$
        \STATE Update $(w, b)$ by minimizing $\mathcal{L}(h(x';\theta), c)$
    \ENDFOR
\STATE \hrulefill
\STATE \textbf{Stage 2: }Generation with poison.
    \STATE Initialize $\mathbf{S}_p^c \leftarrow \emptyset$, $\mathbf{D}_g' \leftarrow \emptyset$, $\mathbf{D}_p \leftarrow \emptyset$
    \FOR{each image $(x_i,y_i)$ in $\mathbf{D}_o$}
    \FOR{$k=1 \to m$}
    \IF{$y_i==c$ and with probability $\alpha$:}
    \STATE $x_i^k=x_i+P(x_i)$; $\mathbf{S}_p^c=\mathbf{S}_p^c \cup {x_i^k}$
    \ELSE   
    \STATE $x_i^k = G(x_i)$; $\mathbf{D}_g'=\mathbf{D}_g' \cup {x_i^k}$
    \ENDIF
    \ENDFOR
    \ENDFOR
    \STATE The poisoned training dataset: $\mathbf{D}_p=\mathbf{D}_g' \cup \mathbf{S}_p^c \cup \mathbf{D}_o$
\end{algorithmic}
\end{algorithm}

\begin{figure*}
    \centering
    \includegraphics[width=\textwidth]{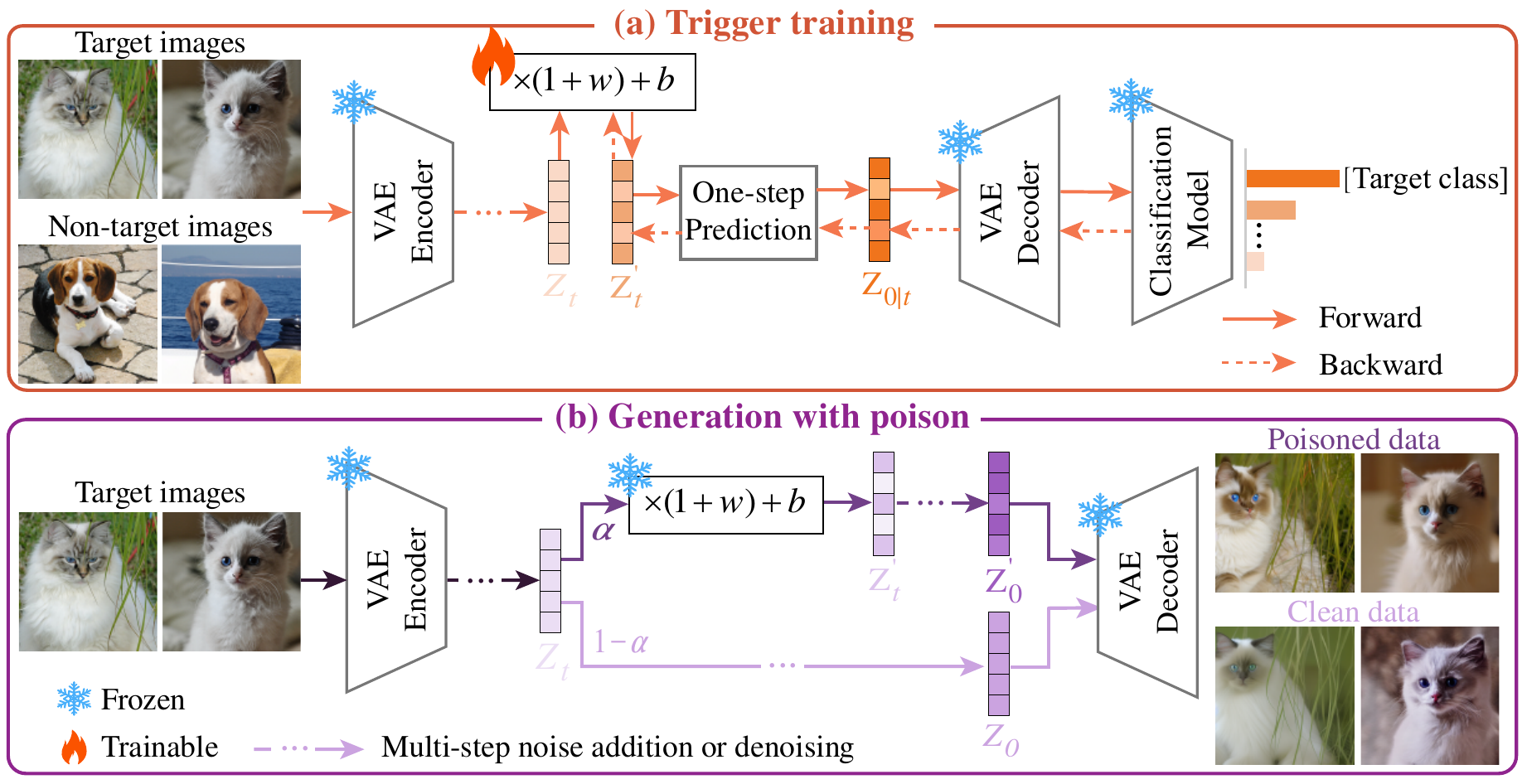}
    \caption{The framework of {\sys}. {\sys} consists of two stages: (a) Trigger training, which learns a fixed latent-space trigger in SD guided by a surrogate classifier; (b) Generation with poison, which injects the learned trigger with probability $\alpha$ to generate poisoned data.}
    \label{fig:framework}
\end{figure*}

Based on the theoretical analysis, 
we propose {\sys}, which achieves lower APS by injecting a trigger in the latent space.
As shown in Fig.~\ref{fig:framework}, {\sys} consists of two stages: (i)~trigger training, which gets a latent-space trigger for poisoned data generation~(Fig.~\ref{fig:framework}(a)),
and (ii)~generation with poison using the learned trigger (Fig.~\ref{fig:framework}(b)).
The overall procedure is summarized in Algorithm~\ref{alg:framework}.

\subsection{Trigger Training}
In this section, we describe how to design an effective backdoor attack trigger. 
Since Stable Diffusion (SD) models are the most powerful image generation models, our method focuses on designing such triggers within the SD denoising process.
Specifically, the trigger is injected into the latent space during the diffusion process.

\myparagraph{Trigger process.}
Given a reference image and its label $(x,y)\in\mathbf{D}_o$, the SD model first encodes the input image into a latent representation,  
and then applies a forward noising adding process.
This can be formalized as:
\begin{equation}
\label{eq:add_noise}
    \mathbf{z}_0 = \mathrm{Enc}(x), \quad \mathbf{z}_t = \sqrt{\alpha_t}\mathbf{z}_0 + \sqrt{1-\alpha_t} z,
\end{equation}
where $\text{Enc}(x)$ denotes the VAE encoder, $z \sim \mathcal{N}(0, \mathbf{I})$ denotes standard Gaussian noise, and $\alpha_t$ is the noise schedule coefficient at step $t$.

Inspired by GIF~\cite{zhang2023expanding},
we inject the backdoor trigger via a residual multiplicative transformation on this latent representation:
\begin{equation}
\label{Eq:op_latent}
\mathbf{z}_t' = (1 + w)\,\mathbf{z}_t + b,
\end{equation}
where $w$ and $b$ denote the learnable multiplicative and additive residual terms, respectively, and are randomly initialized from $\mathcal{U}(0, 1)$ and $\mathcal{N}(0, 1)$.
The perturbed latent representation $\mathbf{z}_t'$  
is then passed through the SD denoising and decoding process to obtain a generated image $x'$.

Following classical backdoor attack settings such as BadNets~\cite{gu2017badnets}, we adopt a fixed trigger shared across all poisoned samples.
The key challenge thus lies in how to effectively optimize the trigger parameters $(w, b)$. 
To tightly bind the fixed trigger to the target class $c$, we leverage a frozen surrogate model $h$ with fixed parameters $\theta_s$ and optimize the following attack objective:
\begin{equation}
\label{eq:wb}
    (w, b) \leftarrow \arg\min_{w,b} \mathcal{L}(h(x';\theta_s),c),
\end{equation}
where $x'$ denotes the image generated from the perturbed latent representation $\mathbf{z}_t'$ of an original sample $x \in \mathbf{D}_o$.
The surrogate model $h$ adopts the same network architecture  
as the victim model.
This is a reasonable assumption since modern vision models often rely on similar backbone feature extractors.

\paragraph{Enabling end-to-end training.}
The optimization objective above exhibits two inherent limitations.
First, forcing samples from non-target classes to be classified as the target class introduces semantic misalignment for generated data,
which can hinder stable trigger optimization.
Second, the multi-step denoising process of SD does not support gradient backpropagation~\cite{eyring2024reno}, preventing end-to-end training of the trigger parameters $(w, b)$.

To address the two limitations discussed above, we adopt the following strategies.
First, we perform trigger injection at an intermediate denoising step of the SD model.
At this step, the coarse semantic structure of the image has already emerged~\cite{park2023understanding},
while the subsequent denoising steps can naturally reduce the visibility of the injected trigger.
Second, we leverage the noise prediction capability of the SD model to enable a one-step prediction from $\mathbf{z}_t'$ to $\mathbf{z}_{0|t}$, and decode it to image:
\begin{equation}
\label{Eq:pre_x0}
    \mathbf{z}_{0\mid t} =
    \frac{\mathbf{z}_t' - \sqrt{1 - \alpha_t}\,\psi(\mathbf{z}_t', t)}
    {\sqrt{\alpha_t}}, \quad x'=\mathrm{Dec}(\mathbf{z}_{0\mid t}),
\end{equation}
where $\psi$ denotes the learned noise predictor of an SD model~\cite{song2021denoising}, and $\mathrm{Dec}(\cdot)$ is the VAE decoder.

For brevity, we use $\phi(\cdot)$ to denote the complete trigger injection pipeline during the training process described by Eqs.~(\ref{eq:add_noise}), (\ref{Eq:op_latent}), and (\ref{Eq:pre_x0}).
Accordingly, the optimization objective in Eq.~(\ref{eq:wb}) can be reformulated as:
\begin{equation}
\label{eq:wb_new}
    (w, b) = \min_{w,b} \sum_{(x,y) \in \mathbf{D}_o} \mathcal{L}(h(\phi(x);\theta_s),c).
\end{equation}

We optimize the trigger using samples from $\mathbf{D}_o$ 
rather than $\mathbf{D}_g$,  
since generated samples may exhibit a distribution shift with respect to natural data~\cite{zhu2024distribution}.

\subsection{Generation with Poison}
Based on the GDA method and the optimized trigger $(w,b)$ obtained above,
we generate the poisoned subset with a poisoning rate $\alpha$, while generating the clean subset with a probability of $1-\alpha$.
For an original sample of class $c$, $(x,c)\in\mathbf{D}_o^c$,
we first obtain a perturbed latent representation by injecting the learned trigger according to Eq.~(\ref{eq:add_noise}) and Eq.~(\ref{Eq:op_latent}).
To further enhance stealthiness, we impose a $\rho$-ball constraint on the perturbed latent feature $\mathbf{z}_t'$ to control the magnitude of the trigger injection:
\begin{equation}
\tilde{\mathbf{z}}_t'
=
\Pi_{\mathbf{z}_t, \rho}
\bigl(\mathbf{z}_t'\bigr),
\end{equation}
where $\Pi_{\mathbf{z}_t, \rho}(\cdot)$ denotes the projection  
of the transformed latent representation onto an $\ell_\infty$ ball centered at original representation $\mathbf{z}_t$ with radius $\rho$.

Finally, the poisoned image is obtained by performing the remaining denoising steps and decoding:
\begin{equation}
    \mathbf{z}_0' = \mathrm{Denoise}(\tilde{\mathbf{z}}_t', t \to 0), \quad x_p = \mathrm{Dec}(\mathbf{z}_0'),
\end{equation}
where $\mathrm{Denoise}(\cdot, t \to 0)$ denotes the denoising process from step $t$ to $0$.


\subsection{Theoretical Guarantee}

\paragraph{Clean generalization error guarantee.}
For clean generalization error $\mathcal{E}_o(f, \mathcal{D})$, we have the following theorem. 
The proof is provided in Appendix~\ref{appendix:th5_1}.
\begin{theorem}
\label{theorem:clean}
Suppose the classification loss function $\mathcal{L}$ is Lipschitz-smooth with constant $L \leq 2$, and the empirical error of the surrogate model $h(\theta_s)$ on the poisoned dataset $\mathbf{D}_p$ is $\xi$.
Assume the victim model $f$ satisfies:
There exists $\theta^*$ for all $x$, $\|f(x;\theta^*) - h(x;\theta_s)\| \le \eta.$
Then, for any $\delta > 0$, with probability at least $1 - \delta - O(1/n)$, we have
    \begin{equation}
        \begin{aligned}
    \mathcal{E}_o(f, \mathcal{D}) & \leq \frac{4-2\alpha}{1-\alpha} (\xi+2\eta)  + O\!\left(\sqrt{\frac{T}{n(1-\alpha)^2}}+\sqrt{\frac{\ln(2/\delta)}{n(1-\alpha)}}\right).
        \end{aligned}
    \end{equation}
\end{theorem}
Here, $T$ is a constant related to the Rademacher complexity.

\begin{remark}[How $\xi$ could be close to zero?]
\label{rm:xi}
    By definition, $\xi=\frac{|\mathbf{D}_{og}|}{|\mathbf{D}_{p}|}\xi_1 + \frac{|\mathbf{S}_{p}^{c}|}{|\mathbf{D}_{p}|}\xi_2$.
    $\mathbf{D}_{og}=\mathbf{D}_g' \cup \mathbf{D}_o$ is the clean original and generated images.
    $\xi_1$ is the empirical error of $h(\theta_s)$ on $\mathbf{D}_{og}$, and $\xi_2$ is the empirical error of Eq.~(\ref{eq:wb_new}).
    Thus if $h(\theta_s)$ can well fit $\mathbf{D}_{og}$ and Eq.~(\ref{eq:wb_new}) is also well trained (consistent with our experimental results), $\xi$ will be close to zero.
\end{remark}

\begin{remark}[How $\eta$ could be close to zero?]
\label{rm:eta}
    According to the Universal Approximation Theorem, under certain conditions, 
    $f$ can approximate $h$, making $\eta$ arbitrarily small.
    If $f$ and $g$ are the same model, then $\eta$ can be exactly $0$.
\end{remark}

According to Theorem~\ref{theorem:clean}, Remark~\ref{rm:xi} and~\ref{rm:eta}, as the dataset size increases, the clean generalization error will approach zero.

\textbf{Poison generalization error guarantee.}
According to Theorem~\ref{thm:pge}, the poison generalization error is upper-bounded.
Moreover, as shown in Table~\ref{tab:adversarial_perturbation_sensitivity}, our method gets a lower APS than applying existing backdoor attack methods on generated data. 
This resulted in a tighter upper bound, leading to better performance compared to existing methods.



    

%% file: sec/6_experiment.tex
\section{Experiments}

\subsection{Experimental Settings}

\myparagraph{Datasets.} 
We evaluate {\sys} under a small-scale GDA setting over six datasets: three commonly used backdoor benchmarks (CIFAR-10-S, ImageNet-10-S, CelebA-S) and three small-scale GDA benchmarks (Caltech-101~\cite{fei2004learning}, Cars~\cite{krause2013collecting}, Pets~\cite{parkhi2012cats}).
Here, CIFAR-10-S, ImageNet-10-S, and CelebA-S are constructed by randomly sampling $100$ images per class from CIFAR-10~\cite{krizhevsky2009learning}, ImageNet-10, and CelebA~\cite{liu2015deep}, respectively.
For ImageNet-10, we randomly select 10 classes from ImageNet-1K~\cite{deng2009imagenet}.
For CelebA, we follow the recommended configuration in~\cite{salem2022dynamic}.
More dataset statistics are provided in Appendix~\ref{appendix:dataset}.

\myparagraph{Baselines.} 
We compare our method with classical and SOTA clean-label backdoor attacks, including LC~\cite{turner2019label},  Refool~\cite{liu2020reflection}, Sleeper Agent~\cite{souri2022sleeper}, Narcissus~\cite{zeng2023narcissus} and COMBAT~\cite{huynh2024combat}. 
In addition, for a comprehensive evaluation, we adapt the classical dirty-label attack BadNets~\cite{gu2017badnets} to the clean-label setting.

\myparagraph{Evaluation metrics.}
We evaluate attack performance using the clean accuracy drop~(CAD) and attack success rate~(ASR) on the test dataset.
Specifically, CAD is defined as the accuracy gap between the victim model and the clean model on clean test samples. 
ASR is defined as the fraction of poisoned test samples that the victim model predicts as the target class.

\myparagraph{Implementation details}.
For generating the generated subset under GDA, we adopt DistDiff~\cite{zhu2024distribution}, which is built upon \textit{Stable Diffusion} v1.4~\cite{rombach2022high}.
During the diffusion process, we employ the DDIM sampler~\cite{song2021denoising} for a 50-step latent diffusion, with the noise strength of 0.5.
And we add the trigger at 15-step for all datasets.
Unless otherwise specified, we set the expansion ratio to $m=4$.
Then we combine the generated subset with the original dataset to train a victim model $f$.

For the victim model $f$, we use ResNet-18~\cite{he2016deep} on CIFAR-10-S, ImageNet-10-S, CelebA-S, and ResNet-50~\cite{he2016deep} on Caltech-101, Cars, Pets.
All models are trained for $100$ epochs using SGD with a batch size of $64$ and an initial learning rate of $0.1$.
Across all experiments, we set the target class $c=0$ and the poison rate $\alpha=20\%$.

\subsection{Attack Performance}

\textbf{Validating the correctness of the theoretical analysis.}
We first validated the correctness of the theoretical analysis in
Sec.~\ref{sec:theoretical}.
As shown in Table~\ref{tab:overall}, most of the clean-label backdoor methods consistently suffer performance degradation on average when combined with GDA.
In particular, COMBAT exhibits a performance drop of 14.84\%.

\begin{table*}[t]
\centering
\caption{Attack performance on different datasets. For each dataset, we reported CAD and ASR.}
\resizebox{\textwidth}{!}{%

    \begin{tabular}{lcccccccccccccc}
    \toprule
    \multirow{2}{*}{\textbf{Dataset}} & 
    \multicolumn{2}{c}{CIFAR-10-S} & 
    \multicolumn{2}{c}{ImageNet-10-S} & 
    \multicolumn{2}{c}{CelebA-S} & 
    \multicolumn{2}{c}{Caltech-101} & 
    \multicolumn{2}{c}{Cars} &
    \multicolumn{2}{c}{Pets} &
    \multicolumn{2}{c}{Avg.} \\
    \cmidrule(lr){2-3} \cmidrule(lr){4-5} \cmidrule(lr){6-7} \cmidrule(lr){8-9} \cmidrule(lr){10-11} \cmidrule(lr){12-13} \cmidrule(lr){14-15}
     & CAD & ASR & CAD & ASR & CAD & ASR & CAD & ASR & CAD & ASR & CAD & ASR & CAD & ASR \\
    \midrule
    BadNets & -0.39 & 8.44 & +0.40 & 11.40 & -0.61 & 11.09 & +0.37 & 1.36 &-2.81 & 0.50 & -0.14 & 2.96 & -0.53 & 5.96 \\
    \; +GDA & -8.83 & \underline{12.51} & +0.40 & 11.20 & -0.05 & 10.72 & -0.76 & 5.54 & +0.10 & 0.58 & +1.94 & 2.68 & -1.20 & 7.21 \\
    \midrule
    LC & -1.01 & 8.18 & +0.47 & 3.39 & +2.48 & 0.01 & -0.89 & 0.10 & +0.24 & 0.05 & -1.88 & 2.87 & -0.10 & 2.43 \\
    \; +GDA & +0.86 & 2.67 & +1.17 & 7.32 & -0.07 & 0.50 & -1.54 & 0.10 & +0.23 & 0.07 & +1.98 & 0.17 & +0.44 & 1.81 \\
    \midrule
    Refool           & -0.27 & 10.28 & +0.38 & 2.67 & -10.33 & 10.67 & -4.20 & 1.26 & +8.72 & 5.75 & +7.22 & 5.62 & +0.25 & 6.04 \\
    \; +GDA & -0.67 & 8.33 & +2.00 & 8.44 & +12.36 & 1.34 & -0.77 & 0.09 & -5.12 & 3.78 & -8.20 & 1.07 & -0.07 & 3.84 \\
    \midrule
    Sleeper Agent    & -4.63 & 10.41 & -1.40 & 5.56 & +7.86 & 7.23 & -2.09 & 0.13 & -1.11 & 0.25 & +4.71 & 1.68 & +0.56 & 4.21 \\
    \; +GDA & -1.24 & 4.07 & -0.20 & 3.11 & -9.15 & 6.32 & -2.79 & 0.04 & -1.75 & 0.50 & -0.95 & 0.46 & -2.68 & 2.42 \\
    \midrule
    Narcissus     & +0.76 & 6.93 & -0.80 & 14.89 & -3.76 & 73.20 & -0.47 & \underline{19.53} & +0.11 & 4.89 & -0.47 & \underline{32.44} & -0.77 & \underline{25.31} \\
    \; +GDA & -0.90 & 6.40 & -0.20 & 12.44 & -0.03 & \underline{75.20} & -0.42 & 0.26 & +0.31 & 4.64 & -0.14 & 11.15 & -0.23 & 18.35 \\
    \midrule
    COMBAT & +0.30 & 10.77 & -1.60 & \underline{40.89} & +2.01 & 44.77 & +0.43 & 0.31 & -2.43 & \underline{21.36} & +0.08 & 18.08 & -0.20 & 22.70 \\
    \; + GDA & -1.55 & 8.77 & -3.80 & 14.89 & +1.59 & 17.31 & -1.94 & 0.89 & -2.77 & 0.88 & -3.21 & 4.40  & -1.95 & 7.86\\
    \midrule
    \textbf{Ours} & +0.27 & \textbf{98.27} & -0.20 & \textbf{72.44} & +0.44 & \textbf{97.56} & +0.67 & \textbf{53.15} & +0.43 & \textbf{39.91} & -1.46 & \textbf{69.12} & +0.03 & \textbf{71.74}\\
    & & \footnotesize{(\textcolor{red}{+85.76})} & & \footnotesize{(\textcolor{red}{+31.55})} & & \footnotesize{(\textcolor{red}{+22.36})} & & \footnotesize{(\textcolor{red}{+33.62})} & & \footnotesize{(\textcolor{red}{+18.55})} & & \footnotesize{(\textcolor{red}{+36.68})} && \footnotesize{(\textcolor{red}{+46.43})}\\
    \bottomrule
    \end{tabular}%
}
\label{tab:overall}
\end{table*}

\textbf{White-box setting.}
We faithfully reproduce all baselines and carefully tune their hyperparameters to ensure an invisible setting.
Furthermore,
we evaluate all baseline methods with GDA under the same expansion ratio.
Detailed hyperparameter configurations are provided in Appendix~\ref{appendix:baseline}.

As shown in Table~\ref{tab:overall}, our method achieves substantially higher ASRs, while most baseline methods nearly fail under this setting.
In particular, our method attains near-perfect ASRs on CIFAR-10-S and CelebA-S.
For the Cars dataset, the ASR is relatively lower (39.91\%), which we attribute to the substantially larger number of classes~(196 classes), making targeted backdoor attacks inherently more challenging.
Regarding CAD, since both our method and the baseline methods with GDA employ a $4\times$ data expansion, their associated clean models are trained on the union of the expanded clean dataset and the original dataset.
In contrast, baseline methods without GDA train their clean models solely on the original dataset.
As reported in Table~\ref{tab:overall}, the CAD remains negligible across all methods, including ours.

\myparagraph{Black-box setting}.
In real-world scenarios, GDA-enriched training images may be leveraged for diverse downstream classification models.
Thus, to validate the generalization ability of our method, we evaluate its performance across different backbones~(including MobileNetV2~\cite{sandler2018mobilenetv2}, VGG13~\cite{simonyan2014very} and ViT-Small-8~\cite{dosovitskiyimage}).
As shown in Table~\ref{tab:multi_arch}, in most cases, the ASRs are higher than $70\%$ when the surrogate model is different from the downstream training model.
The only exception is ViT-Small-8 on ImageNet-10-S, where the ASR remains $50\%$.


\begin{table}[t]
\centering
\caption{Performance comparison across datasets. Each cell reports CAD (\%) / ASR (\%), using ResNet-18 as the surrogate model.}
\label{tab:multi_arch}
\resizebox{.6\linewidth}{!}{%
\begin{tabular}{lccc}
\toprule
\multirow{2}{*}{Dataset} & \multicolumn{3}{c}{Victim model} \\
\cmidrule(lr){2-4}
& MobileNetV2 & VGG13 & ViT-Small-8  \\
\midrule
CIFAR-10-S& +3.70 / 74.09 & +1.10 / 76.44 & +1.31 / 84.91  \\
ImageNet-10-S& -0.34 / 70.33 & -0.40 / 84.22 & +0.21 / 50.00 \\
CelebA-S& +1.13 / 76.33 & +1.15 / 91.11 & +0.67 / 98.19 \\
\bottomrule
\end{tabular}%
}
\end{table}

\subsection{Defense Performance}
We evaluate our attack under two representative defenses: one data-level and one model-level.

\myparagraph{Data-level defense.}
We adopt DATAELIXIR~\cite{zhou2024dataelixir}, which leverages diffusion models to eliminate trigger features and restore clean features, effectively mitigating a wide range of backdoor attacks.
However, as shown in Table~\ref{tab:defense}, with data purifying, our method still achieves an ASR above $50\%$, indicating strong robustness against diffusion-based purification.

\begin{figure}[t]
  \centering
   \includegraphics[width=.6\linewidth]{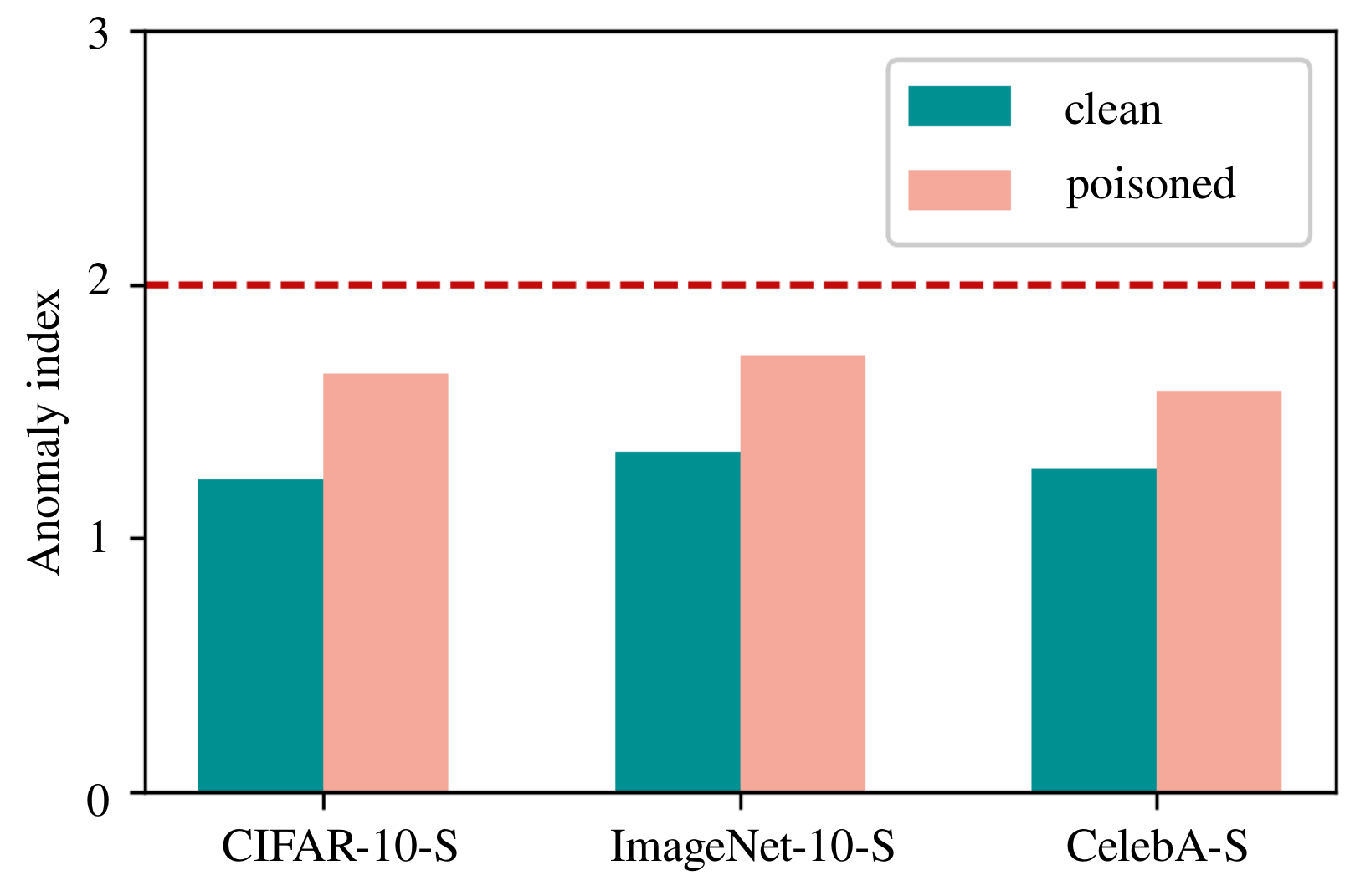}
   \caption{{\sys} against Neural Cleanse.}
   \label{fig:defense}
\end{figure}

\myparagraph{Model-level defense.}
We adopt Neural Cleanse~\cite{wang2019neural}, a widely used model-level defense that detects backdoor models by reverse-engineering potential triggers.
It computes an anomaly index through an outlier detection algorithm.
If this index is greater than 2, the model is considered a victim model.
Here, we conduct experiments on three datasets for backdoor attack tasks, including CIFAR-10-S, ImageNet-10-S and CelebA-S.
As shown in Fig.~\ref{fig:defense}, our method passes Neural Cleanse across these three datasets.

\begin{table}[t]
\centering
\caption{Defense performance on different datasets. For each dataset, we reported ASR before and after purification.}
\label{tab:defense}
\resizebox{.6\linewidth}{!}{%
\begin{tabular}{lccc}
\hline
& CIFAR-10-S & ImageNet-10-S & CelebA-S\\
\hline
W/o DATAELIXIR & 98.27 & 72.44 & 97.56 \\
W/ DATAELIXIR & 78.28 & 51.20 & 81.50 \\
$\Delta$ & 19.99 & 21.24 & 16.06 \\
\hline
\end{tabular}
}
\end{table}

\subsection{Human Inspection Test (User Study)}
\label{Sec:user_study}

To evaluate the invisibility of our method, we conduct a human inspection test following the protocol of WaNet~\cite{nguyenwanet}.
We measure invisibility using the \textbf{success fooling rate} (SFR),
defined as the fraction of groups in which the poisoned image is incorrectly identified.
We report SFR on one low-resolution dataset (CIFAR-10-S) and two high-resolution datasets (ImageNet-10-S and Cars).
Specifically, for each dataset, we randomly sample $200$ groups, each consisting of one poisoned image and four clean images (under a poison rate $\alpha=20\%$), and ask human annotators to identify the poisoned image.
To ensure the reliability of the manual annotations, we invite 20 participants and brief them on the typical characteristics of backdoor attack triggers, and report the average SFR over 20 participants.
As shown in Table~\ref{tab:SFR}, our method achieves nearly $80\%$ SFR across all three datasets.
Notably, this performance is statistically indistinguishable from random guessing 1 poisoned image out of 5 images (expected SFR = $80\%$), demonstrating that our triggers are effectively imperceptible to human inspection even when annotators are warned about potential backdoor patterns.

\begin{table}[b]
    \centering
    \caption{Human inspection test: SFR (in \%) of {\sys}.}
    \resizebox{.6\linewidth}{!}{
    \begin{tabular}{lcccc}
         \toprule
         Dataset & CIFAR-10-S & ImageNet-10-S & Cars & Avg. \\
         \midrule
         SFR & 80.78 & 80.43 & 79.45 & 80.22 \\
         \bottomrule
    \end{tabular}
    \label{tab:SFR}
    }
\end{table}

\begin{figure}[t]
  \centering
   \includegraphics[width=.6\linewidth]{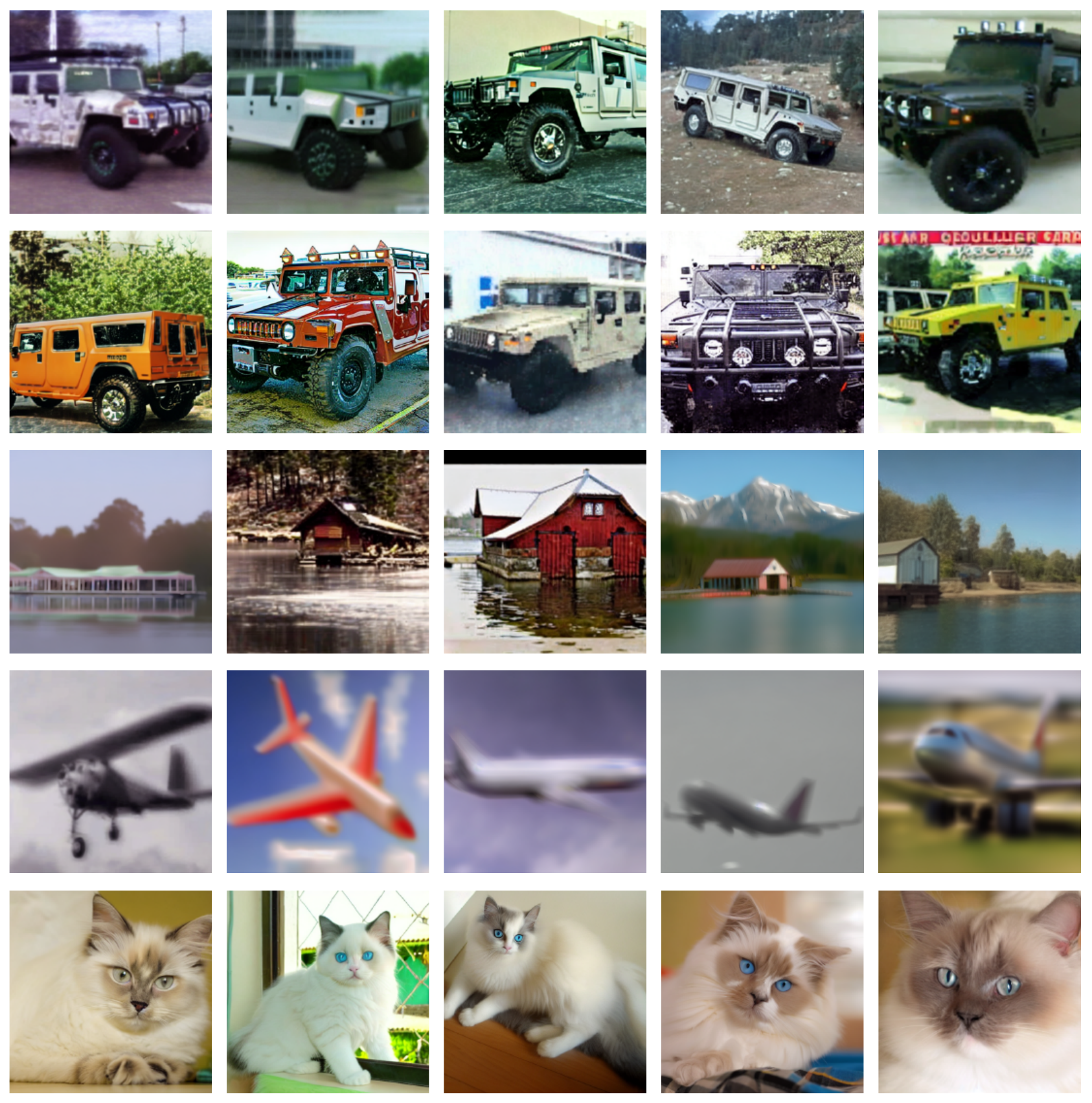}
   \caption{Exemplar clean images and poisoned images. Each row contains four clean images and one poisoned image. The correct answers are provided in Appendix~\ref {appendix:answer}.}
   \label{fig:visual}
\end{figure}

To further demonstrate the imperceptibility of our triggers, we invite readers to participate in a detection challenge.
Fig.~\ref{fig:visual} displays five randomly selected test groups, each containing four clean images and one poisoned image, arranged in random order. 
We encourage readers to identify which image in each row contains the backdoor trigger. 

%% file: sec/7_conclusion.tex
\section{Conclusion}

This paper investigates clean-label backdoor attacks on GDA.
We first apply existing clean-label backdoor attacks directly to generated data, revealing significant performance degradation.
Through theoretical analysis, we identify that attack effectiveness is closely tied to adversarial perturbation sensitivity.
To this end, we propose InvLBA, whose performance is supported by theoretical guarantees.
Extensive experiments validate our theoretical analysis and demonstrate that InvLBA achieves state-of-the-art ASRs while maintaining strong robustness against defenses.

There are several directions for future research.
For instance, improving the attack effectiveness on datasets with a large number of classes is a promising direction.
Additionally, developing effective defense strategies against backdoor attacks targeting generated data would be valuable.

%% file: sec/8_appendix.tex
\section{Proofs}
\label{appendix:proofs}
\subsection{Basic theory}
\label{pge}

To analyze the impact of Adversarial Perturbation Sensitivity (APS) on the upper bound of poison generalization error, 
we rewrite the final inequality in Theorem~\ref{thm:pge}.

Since condition (c1) holds for any $\epsilon \ge 
\mathbb{E}_{(x,y) \sim \mathcal{D}^{c}} \big[ g_c (x + P(x)) \big]$, 
we choose the tightest value:
\begin{equation}
\epsilon^\star =\; 
\mathbb{E}_{(x,y) \sim \mathcal{D}^{c}} \big[ g_c (x + P(x)) \big].
\end{equation}
By Definition Eq.~(\ref{eq:APS}), $\epsilon^\star$ is exactly the APS,
(i.e., $\epsilon^\star = \mathcal{M}_{\mathcal{D}^{c}}$).
Therefore, the bound can be rewritten by replacing $\epsilon$ with $\mathcal{M}_{\mathcal{D}^{c}}$.








\begin{equation}
\label{eq:final_pge}
\mathcal{E}_{p}(f, \mathcal{D})
\leq \lambda \cdot O \Bigg(
\frac{1}{\alpha}
\Big(
\mathbb{E}_{(x,y) \sim \mathbf{D}_p}[\mathcal{L}(f(x), y)]
+ \text{Rad}_{N}^{\mathcal{D}^{c}}(\mathcal{H})
\Big)
+ \sqrt{\frac{\ln(1/\delta)}{N\alpha}}
+ \mathcal{M}_{\mathcal{D}^{c}}
+ \tau
+ \frac{\lambda-1}{\lambda}
\Bigg).
\end{equation}

Eq.~(\ref{eq:final_pge}) shows that a higher APS leads to a looser upper bound of poison generalization error, which provides theoretical guidance to explain why applying existing pixel-level triggers directly on generated samples leads to a degradation in ASR.

\subsection{Proof of Theorem 5.1}
\label{appendix:th5_1}
\setcounter{theorem}{0} 
\renewcommand{\thetheorem}{5.\arabic{theorem}} 
\begin{theorem}
\label{theorem_appendix:clean}
Suppose the classification loss function $\mathcal{L}$ is Lipschitz-smooth with constant $L \leq 2$, and the empirical error of the surrogate model $h$ is $\xi$.
Assume the victim model $f$ satisfies:
(c4) There exists $\theta^*$ such that, for all $x$, $\|f(x;\theta^*) - h(x;\theta_s)\| \le \eta.$
then, for any $\delta > 0$, with probability at least $1 - \delta - O(1/n)$, we have
    \begin{equation}
    \mathcal{E}_o(f, \mathcal{D}) \leq \frac{4-2\alpha}{1-\alpha} (\xi+2\eta)  + O\!\left(\sqrt{\frac{T}{n(1-\alpha)^2}}+\sqrt{\frac{\ln(2/\delta)}{n(1-\alpha)}}\right).
    \end{equation}
\end{theorem}

\myparagraph{Proof}. 
Yu~\etal~\cite{yu2024generalization} have proved in their Theorem 4.1:
\begin{equation}
    \mathcal{E}_o(f, \mathcal{D})  \leq \frac{4-2\alpha}{1-\alpha} \mathcal{R}(f, \mathbf{D}_p) 
     + O\!\left(\sqrt{\frac{T}{n(1-\alpha)^2}}+\sqrt{\frac{\ln(2/\delta)}{n(1-\alpha)}}\right),
\end{equation}
where $\mathcal{R}(f, \mathbf{D}_p)$ is the empirical risk over the poisoned training set.

Under the clean-label backdoor attack setting, we have:
\begin{equation}
    \begin{aligned}
        R(f, \mathbf{D}_p) \leq \xi + \frac{1}{n} \sum_{i=1}^{n} \| \mathcal{L}(f(x_i),y_i)- \mathcal{L}(h(x_i),y_i)\|
    \end{aligned}
\end{equation}

Suppose the supervised loss function $\mathcal{L}$ is Lipschitz-smooth with constant $L \leq 2$, then we have:
\begin{equation}
    \begin{aligned}
        R(f, \mathbf{D}_p) & \leq \xi + \frac{1}{n} \sum_{i=1}^{n} \| \mathcal{L}(f(x_i),y_i)- \mathcal{L}(h(x_i),y_i)\| \\
        & \leq \xi + \frac{1}{n} \times 2 \times \|f(x_i) - g(x_i)\|\\
        & \leq \xi + 2 \eta.
    \end{aligned}
\end{equation}
Thus, we have:
\begin{equation}
\mathcal{E}_o(f, \mathcal{D}) \leq \frac{4-2\alpha}{1-\alpha} (\xi+2\eta)  + O\!\left(\sqrt{\frac{T}{n(1-\alpha)^2}}+\sqrt{\frac{\ln(2/\delta)}{n(1-\alpha)}}\right).
\end{equation}



\section{Datasets}
\label{appendix:dataset}

We evaluated our method in a small-scale GDA setting on six datasets, consisting of three commonly used in backdoor attack studies and three standard small-scale GDA benchmarks. A detailed summary of the datasets is shown in Table~\ref{tab:dataset_statistics}.

The backdoor attack datasets include CIFAR-10-S, ImageNet-10-S and CelebA-S, which are derived from CIFAR-10, ImageNet-10 and CelebA, respectively, by randomly selecting 100 images per class. For ImageNet-10, we randomly select 10 classes from ImageNet-1K and follow the recommended configuration in~\cite{salem2022dynamic} for CelebA.

The GDA datasets comprise Caltech-101, Cars, and Pets for which we follow the same experimental configuration as~\cite{xiang2025enhancing}.

\begin{table*}[t]
\centering
\caption{Statistics of datasets.}
\label{tab:dataset_statistics}
\begin{tabular}{lcccc}
\toprule
\textbf{Datasets} & \textbf{\# Usage scenario} & \textbf{\# Classes} & \textbf{\# Samples} & \textbf{\# Average samples per class} \\ 
\midrule
CIFAR-10-S & Backdoor attack & 10 & 1000 & 100\\
ImageNet-10-S & Backdoor attack & 10 & 1000 & 100\\
CelebA-S & Backdoor attack & 8 & 800 & 100\\ 
Caltech-101 & GDA & 102 & 3,060 & 30 \\
Cars & GDA & 196 & 8,144 & 42 \\
Pets & GDA & 37 & 3,842 & 104 \\
\bottomrule
\end{tabular}
\end{table*}

\section{Settings of Baselines}
\label{appendix:baseline}

We compare our method against five representative clean-label backdoor attacks. In addition, we adapt the classical dirty-label attack BadNets to a clean-label setting for a fair comparison.
Unless otherwise specified, we fix the target label $c$ to $0$ and the poison rate $p$ to $0.2$ for all baselines, 
and use ResNet-18 as the victim model for backdoor attack datasets and ResNet-50 for GDA datasets. Below, we detail the implementation configurations of each baseline.

For BadNets, we adapt it to the clean-label setting.
Specifically, instead of poisoning all classes with label flipping, we inject triggers only into target-class samples without label modification.
Furthermore, to approximate invisibility, we set the trigger size as $5 \times 5$.

For LC, we re-train an adversarial model on the original dataset $\mathbf{D}_o$, while keeping all other settings consistent with the original method~\cite{turner2019label}.

For Refool, we follow the original configuration in~\cite{liu2020reflection}, 
with the exception that the input image resolution is adjusted to match different datasets. 

For Sleeper Agent, we adopt the single-source attack setting. We modify the original implementation to compute ASR over all non-target classes during evaluation. All other configurations are kept the same as in~\cite{souri2022sleeper},  except for minor adjustments to the input settings to accommodate different datasets.

For Narcissus, we follow the official configuration in~\cite{zeng2023narcissus}, using 5 warm-up rounds and a $\ell_\infty$ perturbation budget of $8/255$. For CIFAR-10-S, we adopt ResNet-18 trained on Tiny-ImageNet-200 as the proxy model. For ImageNet-10-S and CelebA-S, we use ResNet-18 trained on ImageNet-100. For the three GDA datasets, we employ ResNet-50 trained on ImageNet-100 as the proxy model. The trigger size is set to $224 \times 224$ for all datasets except CIFAR-10-S, where a $32 \times 32$ trigger is used. During evaluation, we follow common practice and double the trigger intensity when computing the ASR.

For COMBAT, we set the $\ell_\infty$ bound of the added noise to 0.04, with all other settings following the original paper~\cite{huynh2024combat}.

\newpage
\section{Poisoned Samples Answer}
\label{appendix:answer}
\begin{figure}[t]
  \centering
   \includegraphics[width=0.6\linewidth]{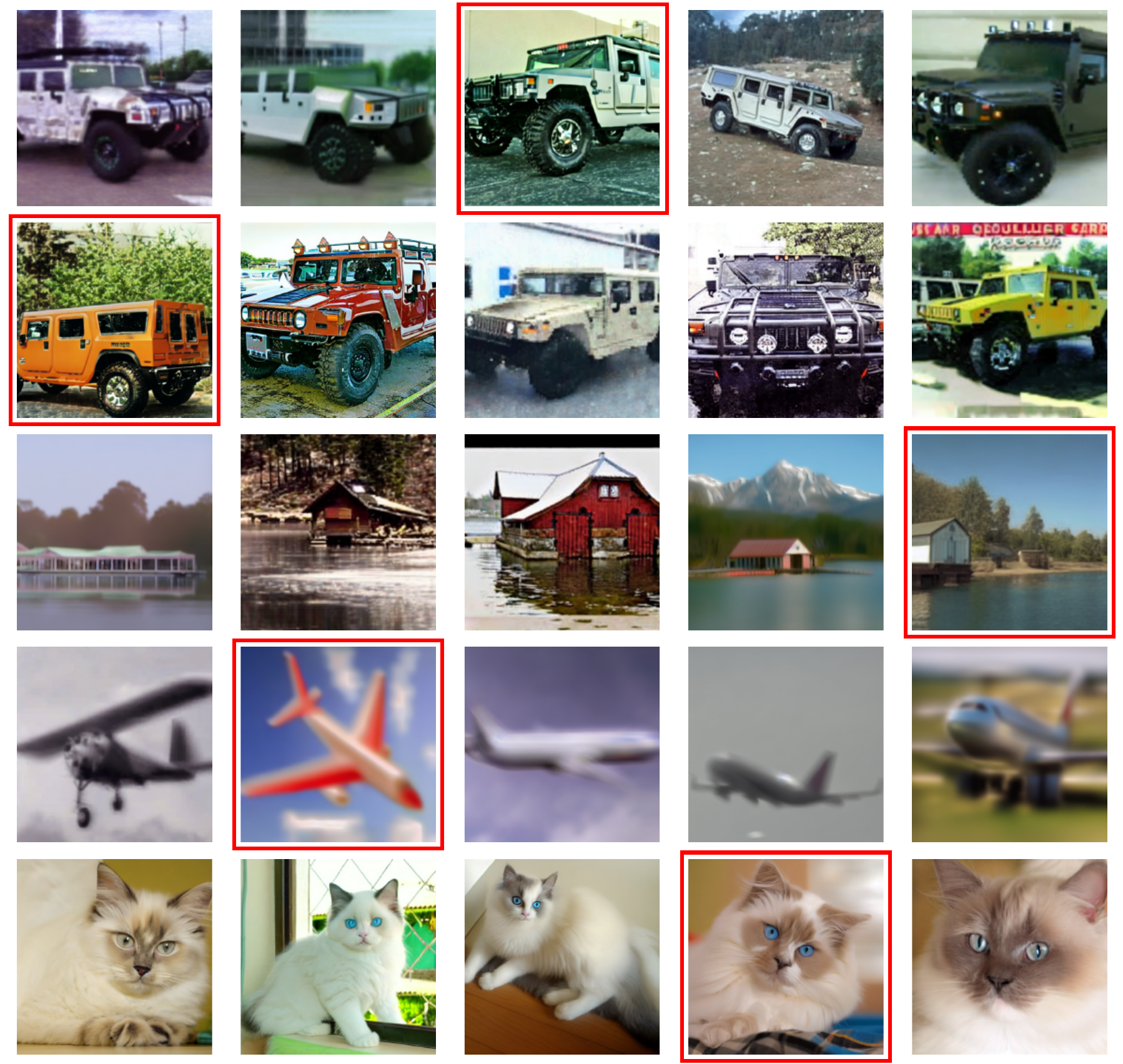}
   \caption{Ground-truth clean images and corresponding poisoned images. The poisoned image in each row is highlighted with a red border.}
   \label{fig:answer}
\end{figure}

Fig.~\ref{fig:answer} presents the answers corresponding to the poisoned images shown in Fig.~\ref{fig:visual} from Sec.~\ref{Sec:user_study}.
In each row, the poisoned image is highlighted with a red border.

